\documentclass[11pt]{article}
\usepackage[letterpaper,margin=1in]{geometry}
\usepackage[T1]{fontenc}
\usepackage{mathptmx}
\usepackage{amsmath,amssymb}
\usepackage{graphicx}
\usepackage{booktabs,array,tabularx}
\usepackage{microtype}
\usepackage{float}
\usepackage{needspace}
\usepackage{fancyhdr}
\usepackage{caption}
\usepackage{authblk}
\usepackage{textcomp}
\usepackage{url}
\usepackage{xurl}
\usepackage[hidelinks,unicode]{hyperref}
\setkeys{Gin}{keepaspectratio}

\hypersetup{
  pdftitle={Better Retrieval, Limited Clustering Gains: A Controlled
Study of Multilingual Company Entity Resolution},
  pdfauthor={Yijiashun Qi; Yuxuan Li; Hanzhe Guo},
  pdfsubject={Multilingual company entity resolution}}
\title{Better Retrieval, Limited Clustering Gains: A Controlled Study of
Multilingual Company Entity Resolution}
\author[1]{Yijiashun Qi\thanks{Corresponding author: elijahqi@umich.edu}}
\author[2]{Yuxuan Li}
\author[1]{Hanzhe Guo}
\affil[1]{University of Michigan, Ann Arbor, USA}
\affil[2]{University of Pennsylvania, Philadelphia, PA, USA}
\date{18 October 2026}
\begin{document}
\maketitle
\thispagestyle{plain}
\begin{abstract}
Company name retrieval and final entity clustering can respond
differently to encoder adaptation. We study this dependency through a
controlled pipeline: train multilingual E5 encoders with matched
positive examples and random or hard negatives, select checkpoints on
development data, retrieve within equal country-specific edge budgets,
and apply unchanged pair classifiers and guarded Union-Find clustering.
Evaluation uses a subsequently collected GLEIF snapshot; a stratified,
assisted human audit and a post-hoc temperature sweep examine label and
parameter sensitivity. On 3,633 names with source-silver labels, adapted
retrieval raises direct pair recall from 53.74\% to 76.98\%, whereas
cluster recall moves from 32.54\% to 33.33\%. Most newly retrieved
positives fail the existing classifier threshold. The study contributes
a controlled comparison and pair-level diagnosis of this loss between
retrieval and confirmation. It shows why evaluating a replacement
retriever requires measuring the resulting entity partition, while
leaving joint matcher retraining and independent gold-label evaluation
for future work.
\end{abstract}


\vspace{1em}
\hypertarget{introduction}{%
\section{Introduction}\label{introduction}}

Multilingual company entity resolution links alternative names across
scripts and sources. It supports knowledge-graph construction, supplier
discovery {[}1{]} and SME analysis {[}2{]}. Missed links fragment
evidence; false merges combine unrelated entities and propagate into
downstream analytics.

Resolution pipelines retrieve candidate pairs, classify them and form
clusters. Better multilingual retrieval may expose matches that an
existing classifier cannot recognize. Transitive clustering can also
turn one incorrect edge into several false relations. Candidate recall
therefore cannot establish final entity quality when a deployed
retriever changes.

Our objective is to measure how retrieval gains survive fixed
confirmation. We compare frozen and adapted multilingual E5 encoders
{[}3{]} under matched positive examples, country eligibility, candidate
budgets and decision rules. We ask whether adaptation improves coverage,
whether hard negatives outperform random negatives, and where new
candidates fail to improve clusters. Legal entities are the target;
shared branding or ownership alone is insufficient.

We contribute an 11-arm comparison under two fixed matchers, pair-level
tracing to complete partitions, and an assisted audit that records
evidence dependence. A revision-stage temperature sweep extends the
analysis. These are empirical contributions using established
components. Sections 3--6 present the workflow, data, experiments and
audit; Sections 7--9 discuss limitations, reproducibility and
conclusions.

\hypertarget{related-work}{%
\section{Related work}\label{related-work}}

JRC-Names-Retrieval benchmarks multilingual name search {[}4{]};
DeepBlocker explores deep blocking and non-neural combinations {[}5{]};
UniBlocker uses contrastive pretraining for transferable blocking
{[}6{]}. These motivate our retrieval choices. We instead measure how
changed candidates survive the same classifier and partition builder,
using established supervised contrastive learning {[}7{]} without
claiming a new loss or state-of-the-art blocking performance.

Ditto fine-tunes language models for pair classification {[}8{]}; our
classifiers remain fixed, isolating retriever replacement without
evaluating joint retraining. RRF {[}9{]} and CSLS {[}10{]} provide
ranking controls, while a capped union supplies a stronger frozen
multi-view baseline.

Short-review embedding studies {[}11{]} and financial reranking
evaluation {[}12{]} relate representations to task outcomes, but do not
study company partitions. Layer-skipping and resilience studies
{[}13{]}, {[}14{]} address efficiency in other workloads; our
equal-candidate budget does not measure those optimizations or serving
cost.

\hypertarget{method-and-workflow}{%
\section{Method and workflow}\label{method-and-workflow}}

Figure 1 summarizes the five steps. Supervision and exposure exclusions
precede encoder adaptation and development-only selection. Each
retrieval arm then supplies an equal candidate budget to frozen
confirmation and guarded clustering. Finally, we evaluate the complete
partition and trace gained pairs to their decisions.

\begin{figure}
\centering
\includegraphics[width=0.95\textwidth,height=\textheight]{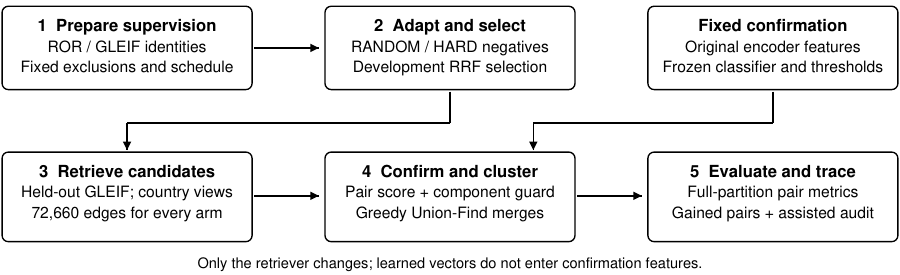}
\caption{Workflow for controlled retriever replacement. Learned vectors
change candidate selection only; confirmation features retain the
original encoder. The temperature sweep repeats adaptation after the
original evaluation.}
\label{fig:workflow}
\end{figure}

\hypertarget{records-and-evaluation-targets}{%
\subsection{Records and evaluation
targets}\label{records-and-evaluation-targets}}

For \(N\) records, \(i,j\in\{1,\ldots,N\}\) index names \(n_i\),
categorical countries \(c_i\) and source identities \(z_i\). Identities
supply labels, never matcher inputs. Eligibility requires \(c_i=c_j\)
without establishing identity; websites and addresses are excluded. The
silver-positive set is
\[
G=\{\{i,j\}:1\leq i<j\leq N,\ z_i=z_j\}.
\]

All retained positives are eligible. A candidate set \(C\) has direct
recall \(|C\cap G|/|G|\). A partition \(\Pi\) comprises nonempty,
disjoint record sets covering all \(N\) records and induces
\[
H(\Pi)=\bigcup_{S\in\Pi}\{\{i,j\}:i,j\in S,\ i<j\}.
\]

Cluster precision is \(|H(\Pi)\cap G|/|H(\Pi)|\) and recall is
\(|H(\Pi)\cap G|/|G|\), both in \([0,1]\) and including transitive
relations. Empty-prediction precision is 1. Identifier equality supplies
silver, not independently adjudicated gold, labels.

\hypertarget{contrastive-encoder-adaptation}{%
\subsection{Contrastive encoder
adaptation}\label{contrastive-encoder-adaptation}}

The encoder maps query-prefixed names to \(v_i\in\mathbb R^d\) (\(d\) is
the positive integer embedding dimension) by masked mean pooling and
unit normalization, giving cosine in \([-1,1]\). A batch samples 16
identities from one source, one natural positive pair per identity and
one explicit negative per anchor. After deduplication, both
positive-pair members form \(Q\). For query \(i\), \(P_i\) contains
other same-identity batch records (\(|P_i|\geq1\)); \(A_i\) contains
these positives and permitted negatives, excluding \(i\). We minimize
\[
\mathcal L=-\frac{1}{|Q|}\sum_{i\in Q}\frac{1}{|P_i|}
\sum_{p\in P_i}\log\frac{\exp(v_i^\top v_p/\tau)}
{\sum_{a\in A_i}\exp(v_i^\top v_a/\tau)}.
\]

Temperature \(\tau>0\) scales similarities before softmax; the loss
pulls same-identity names together relative to competitors. Lower
\(\tau\) emphasizes high-similarity negatives but can amplify unknown
false negatives. All negatives share source/country and differ in
identity/source group, without normalized native/ASCII collisions.
RANDOM samples uniformly from permitted records; HARD samples uniformly
from the original encoder's top 20 permitted neighbors. Mining is static
and positive schedules are identical.

\hypertarget{candidate-selection-at-equal-volume}{%
\subsection{Candidate selection at equal
volume}\label{candidate-selection-at-equal-volume}}

Native and ASCII views use character TF-IDF with 2--4 grams, sublinear
term frequency and full-pool unlabeled statistics. LEXICAL takes the
larger lexical cosine; dense arms use embedding cosine. CSLS subtracts
each endpoint's mean top-10 within-country cosine from twice their
cosine. RRF sums reciprocal directed view ranks,
\(F(i,j)=\sum_v(60+r_v(i,j))^{-1}\). For a country with \(n_c\) names,
directed ranks are integers from 1 to \(n_c-1\); larger fusion scores
are preferred. Dense, lexical, CSLS and RRF arms rank unordered edges by
\[
s_{\mathrm{edge}}(i,j)=\frac{1}{60+r(i,j)}+\frac{1}{60+r(j,i)}.
\]

Here \(r\in\{1,\ldots,n_c-1\}\) is the arm's directed rank and 60 the
fixed smoothing constant. Each country contributes
\(M_c=\min\{\binom{n_c}{2},20n_c\}\) edges, a nonnegative integer,
totaling 72,660 per arm. UNION prioritizes the best rank across
views/directions, then summed reciprocal view ranks and record indices.
Within-selector contrasts isolate adaptation; between-selector contrasts
also change retrieval policy.

\hypertarget{pair-classification-and-guarded-union-find}{%
\subsection{Pair classification and guarded
Union-Find}\label{pair-classification-and-guarded-union-find}}

Frozen GLOBAL-HARD uses histogram gradient boosting with 18 features:
native/ASCII similarities, length/script agreement, original E5 cosine,
CSLS, numeric differences and global token summaries. LOCAL-HARD adds
four neighborhood summaries. Features retain the original encoder;
learned vectors change candidates only. Scores \(s_{ij}\in[0,1]\) are
not demonstrated calibrated probabilities.

Greedy constrained agglomeration uses Union-Find, starting from
singleton names. Process edges with \(s_{ab}\geq\theta\) in descending
score order, breaking ties by record indices. Ignore edges already
inside one component. Otherwise merge current components \(A,B\) iff
\[
\mathrm{merge}(A,B)\iff s_{ab}\geq\theta\ \land\
\mathrm{compatible}(A,B)\ \land\
\min_{i\in A,j\in B}s_{ij}\geq\gamma.
\]

Compatibility requires country agreement for every cross-pair. Both
thresholds lie in \([0,1]\): \(\theta=0.4\) admits the anchor;
\(\gamma=0.05\) blocks any strongly conflicting cross-pair. Missing
scores are computed and cached. Implied relations need not exceed 0.4,
but all within-component scores pass 0.05. Guard \(\gamma\) and
temperature \(\tau\) are distinct parameters.

Union-Find maintains equivalence classes without specifying cluster
count. The cross-pair guard limits erroneous bridge propagation and
yields auditable decisions, matching the task's pair evidence rather
than a centroid objective. The greedy procedure can depend on edge order
and is not globally optimal; cross-pair checking can become expensive
for large components.

\hypertarget{experimental-design}{%
\section{Experimental design}\label{experimental-design}}

\hypertarget{data-provenance-and-coverage}{%
\subsection{Data provenance and
coverage}\label{data-provenance-and-coverage}}

Training uses 28,266 ROR and 5,461 GLEIF names; an observed 4,116-name
holdout becomes development data. ROR covers research organizations
{[}15{]}; GLEIF supplies LEIs and structured legal-name variants
{[}16{]}. Persistent identifiers, name types and saved responses make
labels traceable, but do not guarantee alias correctness or strict
commercial-company coverage.

Original encoders, policies and acquisition rules were frozen before
collecting 39 unused complete 200-record GLEIF pages, timestamped 20
September 2026, 16:00 UTC. We sampled four pages each for DE, FR, ES,
GB, JP, CN, NO, HK and TH, and three for TW, without replacement from
bounded country windows. All requests succeeded without
outcome-dependent expansion. The 7,800 records yield 2,880 identities,
3,633 names and 882 silver-positive pairs.

We retain active GENERAL entities with ISSUED registrations and no
successors. Connected exclusions remove prior raw exposures, known
LEIs/registration identifiers, normalized native/ASCII collisions,
historical ROR names and predefined numbered-family components. This
reduces leakage without ensuring semantic family separation. All
countries and GLEIF were historically seen; coverage supports
held-out-record comparison, not worldwide or unseen-country
generalization.

Names comprise 2,880 primary, 208 alternative-language, 369 preferred
ASCII and 176 automatic ASCII names. Positive pairs include 691
cross-script and 191 same-script pairs; TH supplies 350, whereas GB
supplies distractors only. Single-name entities remain. Country
imbalance, bounded sampling and source-generated transliterations limit
coverage. The audit in Section 6 examines label reliability and its
evidence dependence.

\hypertarget{training-parameters-and-temperature-control}{%
\subsection{Training parameters and temperature
control}\label{training-parameters-and-temperature-control}}

Both arms initialize pinned multilingual-e5-small (revision
614241f622f53c4eeff9890bdc4f31cfecc418b3). FP32 AdamW runs 400 steps,
200/source, with learning rate \(2\times10^{-5}\), weight decay 0.01,
20-step warmup, linear decay, clipping at 1 and seed 20260920. Two
epochs are 200-step cycles totaling 6,400 positive pairs/arm.
Train/evaluation token limits are 128/512; no training name is
truncated. Thirteen ROR identities lacking permitted negatives are
excluded symmetrically.

Original \(\tau=0.05\) was preset, not test-selected or established
optimal. With normalized vectors, its logit scale \(1/\tau=20\) sharpens
similarity differences. The revision adds RANDOM/HARD runs at
\(\tau\in\{0.02,0.10\}\) with identical saved positive/negative
schedules, optimizer, initialization and seed. Development RRF recall
selects epoch 1 or 2 (earlier ties); original 0.05 checkpoints remain
the reference. This post-hoc sweep reuses the evaluation pool without
selecting a new temperature.

\hypertarget{frozen-confirmation-and-analysis-controls}{%
\subsection{Frozen confirmation and analysis
controls}\label{frozen-confirmation-and-analysis-controls}}

Classifiers use 100 iterations, at most 15 leaves, depth 4, minimum leaf
size 25, L2 regularization 1, no early stopping and seed 20260920.
Historical HARD training weights selected high-cosine negatives
fourfold, positives once, with equal total source weight. Global token
features measure weighted Jaccard overlap, smaller/larger unmatched
fractions and maximum unmatched rarity; local features use 20
original-encoder neighbors, excluding the paired record.

Historical \(\theta=0.4\) targets 95\% development precision;
\(\gamma=0.05\) is the retained guard. Weak development RRF results
motivated pre-acquisition UNION controls alongside the primary RRF
contrast, without external preregistration. Checkpoints are not
test-selected.

The 1,000-draw paired bootstrap conditions on models, data and silver
labels, with units connecting identities, families, pages and compared
partitions. It excludes seed/label uncertainty and multiplicity
adjustment. Sixty checks and independent recounts cover 22 original
partitions; all revision partitions and guards are also checked.

\hypertarget{results}{%
\section{Results}\label{results}}

\hypertarget{retrieval-and-final-clusters}{%
\subsection{Retrieval and final
clusters}\label{retrieval-and-final-clusters}}

Table \ref{tab:main-results} retains all 11 original arms. HARD-UNION
raises direct recall from 474/882 to 679/882, a 23.24-point increase
with page-connected 95\% interval {[}18.36, 29.88{]}. RANDOM-UNION
already retrieves 670/882; HARD's additional nine pairs yield 1.02
points with interval {[}-0.78, 2.00{]}, leaving its independent
advantage uncertain.

\begin{table}[!htbp]
\caption{Original equal-budget results. TP, FP, P and R refer to
GLOBAL-HARD complete partitions; the last column gives LOCAL-HARD
counts. All labels are silver; every arm has 72,660 edges.
}
\label{tab:main-results}
\centering\small
\setlength{\tabcolsep}{6pt}
\begin{tabular}{@{}lrrrrrr@{}}
\toprule
Retriever & Ret. R (\%) & TP & FP & P (\%) & R (\%) & LOCAL TP / FP \\
\midrule
LEXICAL & 23.36 & 153 & 5 & 96.84 & 17.35 & 155 / 4 \\
FROZEN-DENSE & 42.97 & 220 & 12 & 94.83 & 24.94 & 214 / 7 \\
FROZEN-CSLS & 40.59 & 219 & 12 & 94.81 & 24.83 & 213 / 7 \\
FROZEN-RRF & 14.85 & 104 & 6 & 94.55 & 11.79 & 104 / 3 \\
RANDOM-DENSE & 75.62 & 267 & 14 & 95.02 & 30.27 & 259 / 9 \\
HARD-DENSE & 76.76 & 267 & 14 & 95.02 & 30.27 & 259 / 9 \\
RANDOM-RRF & 32.77 & 166 & 6 & 96.51 & 18.82 & 166 / 4 \\
HARD-RRF & 33.33 & 166 & 6 & 96.51 & 18.82 & 166 / 3 \\
FROZEN-UNION & 53.74 & 287 & 12 & 95.99 & 32.54 & 282 / 7 \\
RANDOM-UNION & 75.96 & 294 & 14 & 95.45 & 33.33 & 286 / 9 \\
HARD-UNION & 76.98 & 294 & 14 & 95.45 & 33.33 & 286 / 9 \\
\bottomrule
\end{tabular}
\end{table}

Under GLOBAL-HARD, HARD-UNION adds seven correct and two incorrect
co-cluster pairs. Recall rises from 32.54\% to 33.33\%; precision falls
from 95.99\% to 95.45\%. Intervals for the changes are {[}0.27, 1.67{]}
and {[}-1.88, 0.09{]} percentage points. LOCAL-HARD adds four correct
and two incorrect pairs (recall 31.97\% to 32.43\%, precision 97.58\% to
96.95\%); its recall-change interval includes zero (Table~\ref{tab:main-results}).
RANDOM-UNION and HARD-UNION produce identical partitions under each
matcher. The original RRF contrast improves retrieval by 18.48 points
and cluster recall by 7.03, but its frozen baseline is weaker than
frozen dense and lexical retrieval.

\hypertarget{where-retrieval-gains-stop}{%
\subsection{Where retrieval gains
stop}\label{where-retrieval-gains-stop}}

HARD-UNION gains 208 direct positives and loses three relative to
FROZEN-UNION. GLOBAL-HARD scores 202 gains below 0.4 (97.12\%; median
over all gains 0.0479). Six pass and co-cluster; the seventh additional
correct relation is transitive. LOCAL-HARD admits three gained direct
positives. This locates an immediate threshold bottleneck without
identifying features, training examples or calibration as its unique
cause. Lowering the threshold is not shown safe: new false pairs already
pass it.

Positive-edge-only closure connects 718/882 positives (81.41\%).
Unrestricted connectivity can require false edges and therefore does not
establish attainable perfect clustering. Cross-script coverage increases
from 385/691 to 553/691, and same-script coverage from 89/191 to
126/191. These heuristic script strata and the uneven country mix limit
language-specific interpretation.

\hypertarget{temperature-sensitivity}{%
\subsection{Temperature sensitivity}\label{temperature-sensitivity}}

UNION retrieval recall ranges from 73.47\% to 76.98\% across the three
temperatures (Table \ref{tab:temperature}). The counts report
complete-partition outcomes under each fixed matcher. This post-hoc
sweep does not establish an optimal temperature or multi-seed
robustness.

\begin{table}[!htbp]
\caption{Post-hoc temperature sensitivity with UNION selection and fixed
confirmation. The original 0.05 checkpoints are reused; every arm has
72,660 candidates and 882 silver-positive pairs. }
\label{tab:temperature}
\centering\small
\setlength{\tabcolsep}{6pt}
\begin{tabular}{@{}rlrrr@{}}
\toprule
\(\tau\) & Negatives & Ret. R (\%) & GLOBAL TP / FP & LOCAL TP / FP \\
\midrule
0.02 & RANDOM & 74.83 & 294 / 14 & 286 / 9 \\
0.02 & HARD & 76.98 & 294 / 14 & 286 / 9 \\
0.05 & RANDOM & 75.96 & 294 / 14 & 286 / 9 \\
0.05 & HARD & 76.98 & 294 / 14 & 286 / 9 \\
0.10 & RANDOM & 73.47 & 294 / 12 & 286 / 8 \\
0.10 & HARD & 73.47 & 294 / 13 & 286 / 9 \\
\bottomrule
\end{tabular}
\end{table}

Development-only selection chooses 0.02 RANDOM epoch 1, 0.02 HARD epoch
2, 0.10 RANDOM epoch 2, 0.10 HARD epoch 2. All four added runs complete
400 updates; independent recounts and cross-pair checks validate all
eight new partitions. The original experiment remains unchanged.

\hypertarget{assisted-identity-audit}{%
\section{Assisted identity audit}\label{assisted-identity-audit}}

The 137-pair audit includes all 14 false co-cluster pairs across main
UNION outputs, all three lost positives and up to 30 sampled pairs per
remaining stratum, with recorded inclusion probabilities. Initial inputs
showed names, country and randomized IDs, withholding model
outputs/private linkage. Thirty was a workload cap, not a power
calculation.

On 3 October 2026, one reviewer checked every pair and supporting
evidence with AI assistance. Table \ref{tab:annotation} records 92 SAME,
44 DIFFERENT and one UNSURE judgments with rationale/URL. All 136
resolved judgments agree with silver labels. The 30 sampled gained
positives remain rejected; all 14 false co-cluster pairs, including two
new errors, are judged DIFFERENT.

\begin{table}[!htbp]
\caption{Assisted single-reviewer judgments. LEI-only counts classify
submitted URLs; they do not certify all evidence content.
}
\label{tab:annotation}
\centering\small
\setlength{\tabcolsep}{6pt}
\begin{tabular}{@{}lrrrr@{}}
\toprule
Sampling stratum & Same & Different & Unsure & LEI-only \\
\midrule
Positive, retrieved by both & 30 & 0 & 0 & 24 \\
Positive, newly retrieved & 30 & 0 & 0 & 24 \\
Positive, lost after training & 3 & 0 & 0 & 2 \\
Positive, retrieved by neither & 29 & 0 & 1 & 24 \\
Negative, predicted co-cluster & 0 & 14 & 0 & 8 \\
Other silver-negative pairs & 0 & 30 & 0 & 25 \\
\bottomrule
\end{tabular}
\end{table}

Evidence for 107/137 entries, including 24/30 gained positives, cites
only GLEIF/LEI pages; other URLs need not be independent either. GovMem
studies correlated evidence in agent memory {[}17{]}; here, repeated
registry evidence likewise cannot independently validate registry
labels. One unresolved pair remains UNSURE. Single-reviewer,
source-dependent judgments cannot establish gold accuracy or agreement.
Unequal sampling prevents population-precision estimates; 30 residual
negatives cannot tightly bound hidden matches among 806,941 pairs.

\hypertarget{discussion-and-limitations}{%
\section{Discussion and limitations}\label{discussion-and-limitations}}

Fixed-confirmation results do not establish behavior under joint
retraining. One seed, seen sources/countries, heuristic filtering,
source transliterations and silver labels restrict generalization. The
post-hoc sweep does not resolve optimality or seed variability.
Exhaustive transductive retrieval assumes complete country metadata;
approximate-search latency, missing countries, website/domain evidence,
incremental updates and production SL2 overrides remain untested.

Future work should train logistic-regression or HGB matchers on
out-of-fold candidates generated by the deployment retriever, add
learned cosine features, and compare recalibration separately. A
fixed-candidate factorial experiment could distinguish feature changes
from training-distribution and threshold effects. Thresholds should be
chosen on development data and assessed on a fresh test sample.
Independent dual annotation, broader country coverage, multiple training
seeds and guard-cost measurements would strengthen validity; additional
tuning must treat the present pool as development data.

\hypertarget{reproducibility-and-ai-assistance}{%
\section{Reproducibility and AI
assistance}\label{reproducibility-and-ai-assistance}}

Protocols, acquisition responses, checkpoints, training schedules,
feature matrices, scores, partitions and annotation records are retained
locally. Reproduction requires the saved snapshot because live
pagination changes; a public archive is not yet available. AI tools were
used only to polish implementation code; all research ideas,
methodological choices, analyses and interpretations are the authors'
own. One reviewer verified the assisted annotations, as described in
Section 6. The authors are responsible for the manuscript's accuracy.

\Needspace{9\baselineskip}
\hypertarget{conclusion}{%
\section{Conclusion}\label{conclusion}}

Contrastive adaptation substantially improves retrieval under equal
budgets, while fixed confirmation adds few correct relations and new
errors. Pair tracing locates most lost gains at the classifier
threshold. The study provides a controlled diagnosis for knowledge
integration; it does not establish that jointly retrained systems have
the same limitation. Future evaluation must test confirmation changes
and independent identity labels.

\hypertarget{refs}{}

\end{document}